\documentclass{article} 
\usepackage{iclr2026_conference,times}

\usepackage{amsmath,amsfonts,bm}

\def\eqref#1{equation~\ref{#1}}

\def\1{\bm{1}}

\DeclareMathAlphabet{\mathsfit}{\encodingdefault}{\sfdefault}{m}{sl}
\SetMathAlphabet{\mathsfit}{bold}{\encodingdefault}{\sfdefault}{bx}{n}

\usepackage{hyperref}
\usepackage{url}

\usepackage{graphicx}
\usepackage{subcaption}
\usepackage{amsmath}
\usepackage{booktabs}
\usepackage{multirow}
\usepackage{float}
\usepackage{xcolor}
\usepackage{colortbl}
\usepackage{wrapfig}
\usepackage{fontawesome5}

\newcommand{\ourbench}{WildASR}

\title{Back to Basics: Revisiting ASR in the Age of Voice Agents}

\author{\parbox{13cm}{Geeyang Tay\textsuperscript{$\dagger$}, Wentao Ma\textsuperscript{$\dagger$}\thanks{Corresponding to: wentao@boson.ai. $\dagger$ Equal contribution.} , Jaewon Lee, Yuzhi Tang, Daniel Lee, Weisu Yin, \\ Dongming Shen, Silin Meng, Yi Zhu, Mu Li, Alex Smola} \vspace{1.5mm}\\
Boson AI
}

\newcommand{\err}[1]{\textbf{#1}} 
\newcommand{\mono}[1]{\texttt{#1}}
\iclrfinalcopy 

\renewcommand{\headrulewidth}{0pt}

\begin{document}

\maketitle

\begin{abstract}
Automatic speech recognition (ASR) systems have achieved near-human accuracy on curated benchmarks, yet still fail in real-world voice agents under conditions that current evaluations do not systematically cover. Without diagnostic tools that isolate specific failure factors, practitioners cannot anticipate which conditions, in which languages, will cause what degree of degradation. We introduce \ourbench{}, a multilingual (four-language) diagnostic benchmark sourced entirely from real human speech that factorizes ASR robustness along three axes: environmental degradation, demographic shift, and linguistic diversity. Evaluating seven widely used ASR systems, we find severe and uneven performance degradation, and model robustness does not transfer across languages or conditions. Critically, models often hallucinate plausible but unspoken content under partial or degraded inputs, creating concrete safety risks for downstream agent behavior. Our results demonstrate that targeted, factor-isolated evaluation is essential for understanding and improving ASR reliability in production systems. Besides the benchmark itself, we also present three analytical tools that practitioners can use to guide deployment decisions.

\includegraphics[height=2.5ex]{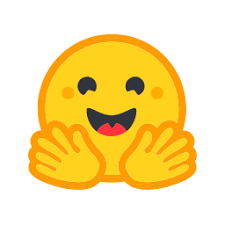} \url{https://huggingface.co/datasets/bosonai/WildASR} \\
\includegraphics[height=2.5ex]{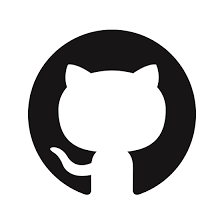} \url{https://github.com/boson-ai/WildASR-public}

\end{abstract}

\section{Introduction}
\label{sec:introduction}

The field of Automatic Speech Recognition (ASR) has witnessed a decade of unprecedented progress, driven largely by the scaling of neural architectures and the availability of massive datasets. The declaration of ``human parity'' by~\citep{amodei_2016_deepspeech2,xiong_2017_asrHumanParity} marked a pivotal moment, and this progress has been further accelerated by~\citep{zhang_2020_audioSSL,radford_2022_whisper,pratap_2023_mms} which leverage hundreds of thousands of hours of web-scraped audio to achieve remarkable performance across diverse languages.
Contemporary systems now routinely obtain word error rates (WER) lower than $5\%$ on curated benchmarks~\citep{panayotov2015librispeech,ardila2020commonvoice}. This rapid advancement raises the question: \emph{Is multilingual ASR a solved problem?}

\begin{figure}[h]
    \centering
    \includegraphics[width=\linewidth]{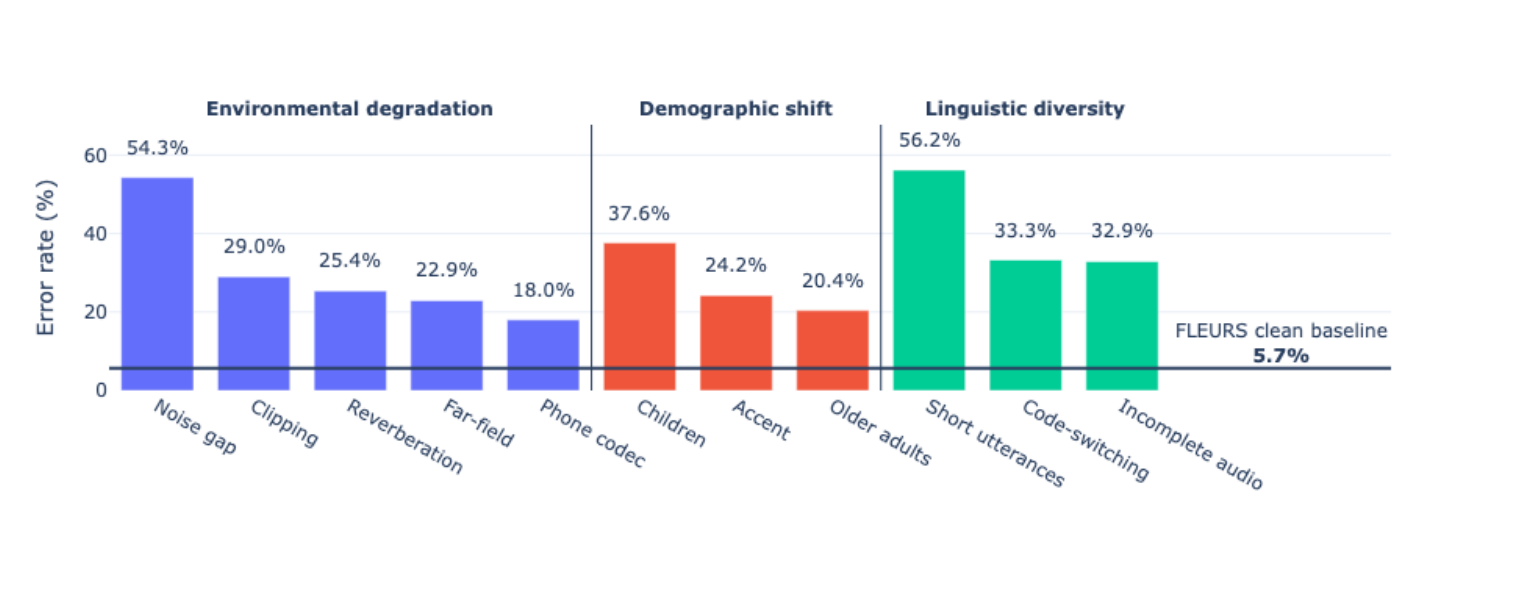}
    \vspace{-8ex}
    \caption{\textbf{Multilingual ASR robustness under real-world distribution shifts in \ourbench{}.} We evaluate seven ASR systems across four languages and aggregate performance over three OOD dimensions. The horizontal line denotes the in-distribution clean-set model-average reference (5.7\%), defined as the average error rate on the FLEURS test set across all models and languages. The sharp and uneven degradation across OOD conditions shows that human-parity performance on in-distribution data does not reliably transfer to real-world settings.}
\label{fig:overall_env}
\end{figure}

Voice agents, AI systems capable of engaging in spoken dialogue with users, have been rapidly proliferating in the past few years~\citep{shi2025voila, zeng2024glm, arora2025stream}. As voice emerges as a dominant interface modality, these agents must contend with a wide spectrum of out-of-distribution (OOD) conditions: telephony compression, overlapping speech, regional accents, disfluencies, and code-switching. For the well performed ASR system, when they are deployed in real-world voice agents, failures still occur~\citep{chen2024voicebench, jain2025voiceagentbench, xu2025voiceagenteval}.

Moreover, voice agents do not merely use ASR outputs as passive transcription, but rely on them to trigger downstream tools, retrieve context, and execute actions. Under OOD conditions, transcription errors are not merely cosmetic, as can be seen in Figure~\ref{fig:overall_env}. Yet existing ASR evaluations predominantly test on in-domain data and report aggregate word error rate (WER)~\citep{panayotov2015librispeech,ardila2020commonvoice,shah2024speech,wang2025audiobench, sakshi2024mmaumultitask}, obscuring which specific acoustic or linguistic factors drive failures. As a result, current ASR benchmarks cannot answer whether robustness to one perturbation transfers across languages, environments, or conversational settings. This creates a \emph{diagnostic gap}: practitioners have no systematic way to identify \emph{where} (environment), \emph{who} (demographics), and \emph{what} (linguistic phenomena) drives ASR failures in their specific deployment. To close this diagnostic gap, we introduce \textbf{\ourbench{}}, a multilingual (four-language) benchmark that provides systematic, factor-isolated evaluation of ASR robustness under real-world OOD conditions. Our contributions are threefold:

\begin{itemize}
    \item \textbf{\ourbench{}: a diagnostic benchmark for real-world OOD shifts} We introduce a multilingual (four-language), multi-dimensional benchmark whose source audio comes entirely from real human speech rather than TTS-generated data. To systematically isolate failure modes, we decompose robustness into three axes: Environmental Degradation (the where), Demographic Shift (the who), and Linguistic Diversity (the what).
    \item \faRulerCombined \textbf{ Thorough evaluation under a unified protocol} We benchmark seven state-of-the-art systems (including proprietary and open-source models) under a unified protocol. We report both standard metrics and factor-isolated degradations, revealing that robustness does not transfer reliably, and performance rankings fluctuate wildly across languages.
    \item \faMedapps \textbf{ Diagnostic analyses as deployment decision tools} Moving beyond average WER, we characterize specific deployment risks and present analytical tools that practitioners can directly apply: a P90 Elbow analysis to identify instability thresholds under increasing distortion, prompt sensitivity profiling to quantify variance from instruction phrasing, and hallucination error rate to expose semantic fabrications in linguistic edge cases.
\end{itemize}

\section{Related work}
\label{sec:related}

\paragraph{ASR} 
Modern ASR systems have advanced rapidly due to self-supervised learning and large-scale multilingual training. Models ~\citep{baevski2020wav2vec2,gulati2020conformer,radford_2022_whisper} have achieved near-human accuracy on widely used benchmarks ~\citep{panayotov2015librispeech,hernandez2018tedlium3,ardila2020commonvoice,pratap2020mls,pratap_2023_mms,fleurs2022arxiv}. These datasets have played a critical role in driving progress by standardizing evaluation and enabling fair comparison.

However, these ASR benchmarks largely reflect in-distribution conditions, resulting in saturated performance and limited insight into system behavior under realistic deployment shifts~\citep{Koenecke2024carelesswhisper,Barański2025investigation,frieske2024hallucinationsneuralautomaticspeech}.
To address this gap, several works study ASR robustness under specific perturbations such as additive noise, reverberation, accented speech, or domain mismatch~\citep{shah2024speech,wang2025contextasr}, demonstrating substantial degradation under adverse conditions and motivating robustness-oriented training.
While valuable, these evaluations typically focus on a limited set of languages or datasets and often rely on TTS-generated speech to construct test samples. However, synthetic speech lacks the authentic paralinguistic phenomena present in real human recordings~\citep{liao2025nvspeech, li2024spontaneous}, such as hesitations, disfluencies, and unstable articulation, and can substantially underestimate failure rates (see Section~\ref{sec:discussion} for empirical evidence). Preserving real human speech sources is therefore critical for valid robustness evaluation.
In contrast, our \ourbench{} sources all audio from real human speech and applies controlled augmentations to enable factorized evaluation across multiple perturbation axes and languages, enabling systematic analysis of ASR failure modes and robustness trade-offs.

\paragraph{AudioLLM} Recent work has explored integrating speech understanding with large language models~\citep{sun2024salmonn, qwen2023audio, ghosh2024gama, rubenstein2023audiopalm, huang2023audiogpt, scribe-v1, gpt4o-transcribe, qwen2-audio, nova-2}, giving rise to AudioLLMs that combine pretrained audio encoders with text-centric LLM backbones for unified speech recognition, translation, and audio reasoning, or even with other modalities~\citep{comanici_2025_gemini25, gemini-pro-3}. Parallel efforts have explored end-to-end speech-to-speech (S2S) systems~\citep{defossez2024moshi, google2025geminilive, openai2025gptrealtime, roy2026personaplex, wu2025stepaudio2technicalreport}, which reduce latency and preserve paralinguistic cues. 

To evaluate such models, benchmarks ~\citep{wang2025audiobench,chen2024voicebench,sakshi2024mmaumultitask} emphasize multimodal audio understanding and reasoning rather than transcription accuracy alone. More efforts ~\citep{cheng2025ahabench,liu2025vocalbenchDF,zhang2025wildspeech,koudounas2025hallucinationbenchmarkspeechfoundation, zhang2024benchmarking, liu2025vocalbench} try to highlight hallucination and instability in audio-language systems\citep{sun2024salmonn, kuan2024largeaudiolanguagemodelstruly, atwany2025lost, wang25b_interspeech}. However, these benchmarks often evaluate task success, implicitly assuming that upstream ASR outputs are sufficiently reliable. In contrast, our \ourbench{} focuses specifically on the trustworthiness of ASR as a foundational component in such systems. By exposing substantial transcription failures under realistic conditions, \ourbench{} highlights a critical gap between benchmark ASR performance and the reliability required for safe downstream decision-making.

\vspace{-2ex}
\section{\ourbench{}}
\label{sec:method}

Real-world voice agents encounter a long tail of acoustic and linguistic conditions that curated benchmarks rarely cover, and these conditions can trigger not just higher error rates but outright hallucinations. Rather than optimizing for average-case accuracy, we construct a diagnostic benchmark that (i) reflects real voice-chat usage, (ii) isolates concrete OOD factors, and (iii) enables per-factor analysis. We organize these factors into three dimensions: \emph{environmental degradation (the where), demographic shift (the who), and linguistic diversity (the what)}.

To operationalize these failure modes, we construct \textbf{\ourbench{}}. We first describe our curation pipeline, then detail each dimension. An overview is presented in Table~\ref{tab:benchmark-construction}.

\vspace{-2ex}
\subsection{Curation pipeline}
\label{subsec:curation}

The design of \ourbench{} follows a \emph{real source, controlled perturbation} principle: all source audio originates from real human speech to preserve authentic paralinguistic phenomena (e.g., hesitations, disfluencies, and articulatory variation) that TTS systems fail to reproduce; controlled augmentations are then applied post-hoc to isolate specific acoustic factors without introducing synthetic artifacts. The benchmark covers four languages: English (EN), Chinese (ZH), Japanese (JA), and Korean (KO), with three distinct data splits corresponding to our three OOD dimensions. 

\noindent The curation pipeline consists of seven stages: \textbf{DC} (Data Collection), \textbf{SF} (Speaker Filtering), \textbf{QF} (Quality Filtering), \textbf{NR} (Audio Normalization), \textbf{AA} (Acoustic Augmentation), \textbf{MT} (Manual Truncation \& Transcript Alignment), and \textbf{MV} (Manual Verification). Not all steps apply to every subset; the rightmost column of Table~\ref{tab:benchmark-construction} indicates which steps were applied to each subcategory. Detailed descriptions of each step are provided in Appendix~\ref{sec:curation_details}.

\vspace{-2ex}
\subsection{Environmental degradation}
\label{subsec:environmental_degradation}

Voice agents operate on user-generated audio recorded in uncontrolled conditions that are often far from the distributions represented in standard ASR training and evaluation. To isolate environment-driven acoustic shifts while keeping the linguistic content fixed, we apply five controlled, transcript-preserving augmentations to each utterance. 

\noindent \textbf{Reverberation} Reverberation is one of the most common factors that reduces indoor audio quality. We simulate room acoustics using the image-source method~\citep{pyroomacoustic}, which introduces temporal smearing from reflections. We parameterize severity by the reverberation time $\mathrm{RT}_{60}$ (i.e., the time for the sound energy to decay by 60\,dB). To be specific, we vary $\mathrm{RT}_{60}$ across three distinct levels $(0.4/0.8/1.6s)$ to cover mild to strong reverberation.

\noindent \textbf{Far-field}  Distinct from simple reverberation (which relies on room absorption), far-field audio is characterized by a low direct-to-reverberant ratio~\citep{haebumbach_2020_farFieldASR}. 
It creates a smearing effect where reflections overwhelm the direct path, severely degrading the intelligibility of short phonemes (e.g., consonants). To isolate this effect, we fix the room acoustics ($\mathrm{RT}_{60}$) using a simulated room geometry, and vary only the source-microphone distance to $(4/8/16m)$.

\noindent \textbf{Phone Codec} Real-world voice agents frequently encounter narrowband telephone audio rather than wideband, studio-like recordings. To simulate legacy communication channels, we process audio through two standard codecs: GSM (representing classic mobile telephony) and G.711 $\mu$-law (representing standard landline/VoIP infrastructure). Both operations involve downsampling the input to $8$\,kHz, applying the codec's quantization artifacts, and resampling back to $16$\,kHz, testing the model's ability to recover phonemes from band-limited representations.

\noindent \textbf{Noise gap} Hallucinations are often associated with long non-speech spans within an utterance~\citep{Koenecke2024carelesswhisper}. To probe this failure mode, we inject synthetic stationary noise between contiguous speech fragments,  increasing the non-vocal duration while preserving the original lexical content. Specifically, we vary the density and duration of these insertions: $3$ or $5$ gaps of either $0.2$\,s or $0.4$\,s duration, leading to $(N_{\text{gap}},\Delta t)\in\{(3,0.2),(5,0.2),(3,0.4),(5,0.4)\}$. This stresses the model's endpointing mechanisms without introducing confounding linguistic complexity.

\noindent \textbf{Clipping} Clipping occurs when input gain saturates the recording hardware (e.g., loud speech or background music), clamping the waveform against a maximum limit. We model this by setting a per-utterance clipping threshold such that the top $40\%$ of signal amplitude values are flattened, followed by RMS rescaling to recover loudness. This introduces harsh non-linear harmonic distortion that standard noise-robustness techniques often fail to model.

\noindent We establish a high-quality base corpus by sampling utterances from two complementary sources: the  widely adopted FLEURS~\citep{fleurs2022arxiv} test split, which provides read speech, and a few  conversational datasets from MagicData~\citep{zhou_2025_magicdataFullDuplex,magicdata_2024_kcsc,magicdata_2025_jaDuplex}, which captures spontaneous speech. Both sources cover all four target languages. We discard unintelligible samples, and apply five controlled perturbations to enable factor-isolated analysis.

\begin{table*}[t]
\centering
\small
\caption{\textbf{Overview of the proposed \ourbench{}.} Each OOD dimension is decomposed into explicitly defined subcategories. For each subcategory, we report the covered languages, the number of samples per language, the average utterance duration, and the curation steps applied (defined in \S\ref{subsec:curation}). Detailed data sources are listed in Appendix~\ref{sec:data_sources}.}
\vspace{-1ex}
\setlength{\tabcolsep}{4pt}
\renewcommand{\arraystretch}{1.12}
\resizebox{0.9\textwidth}{!}{%
\begin{tabular}{l c c c c}
\toprule
\textbf{Categories} & \textbf{Languages} & \textbf{\#Samples} & \textbf{Avg Duration (s)} & \textbf{Curation Steps} \\
\midrule
\textbf{Environmental degradation}  & & & & \\
\quad Reverberation & EN/ZH/JA/KO & 2841/3735/2850/2046 & 10.0/12.9/12.5/10.4 & DC$\to$QF$\to$NR$\to$AA$\to$MV \\
\quad Far-field & EN/ZH/JA/KO & 2841/3735/2850/2046 & 10.0/13.0/12.5/10.4 & DC$\to$QF$\to$NR$\to$AA$\to$MV \\
\quad Phone codec & EN/ZH/JA/KO & 1894/2490/1900/1364 & 7.5/10.4/10.0/7.9 & DC$\to$QF$\to$NR$\to$AA$\to$MV \\
\quad Noise gap & EN/ZH/JA/KO & 3788/4980/3800/2728 & 8.6/11.6/11.2/9.1 & DC$\to$QF$\to$NR$\to$AA$\to$MV \\
\quad Clipping & EN/ZH/JA/KO & 947/1245/950/682 & 7.5/10.4/10.0/7.9 & DC$\to$QF$\to$NR$\to$AA$\to$MV \\
\midrule
\textbf{Demographic shift} & & & & \\
\quad Children & EN/ZH & 300/1000 & 4.06/3.02 & DC$\to$SF$\to$QF$\to$NR$\to$MV \\
\quad Older adults & EN/ZH & 300/1000 & 5.93/1.95 & DC$\to$SF$\to$QF$\to$NR$\to$MV \\
\quad Accent & EN/ZH & 1000/1000 & 3.48/5.69 & DC$\to$SF$\to$QF$\to$NR$\to$MV \\
\midrule
\textbf{Linguistic diversity} & & & & \\
\quad Short utterances & EN/ZH/JA/KO & 318/367/467/255 & 1.2/0.7/1.1/1.0 & DC$\to$QF$\to$NR$\to$MV \\
\quad Incomplete audio & EN/ZH/JA/KO & 2345/2517/195/396 & 3.9/2.1/1.9/2.6 & DC$\to$QF$\to$NR$\to$MT$\to$MV \\
\quad Code-switching & (EN)+ZH/JA/KO & 700/700/700 & 8.6/11.7/11.5 & DC$\to$QF$\to$NR$\to$MV \\
\bottomrule
\end{tabular}
}
\vspace{-3ex}
\label{tab:benchmark-construction}
\end{table*}

\vspace{-2ex}
\subsection{Demographic shift}
\label{subsec:demographic_shift}

Standard ASR training and evaluation corpora are dominated by working-age adults speaking relatively standard varieties. This mismatch constitutes both a fairness concern and a product risk, particularly salient for two fast-growing use cases: children's education and geriatric care. To bridge this gap, we curate three sub-corpora that represent critical user groups for which current systems often fail.

\noindent \textbf{Children} Recognition of child speech is uniquely challenging due to higher fundamental frequency, irregular prosody, frequent disfluencies, and evolving linguistic patterns that defy adult-trained model assumptions. In \ourbench{}, we source English data from Zenodo's children speech recording~\citep{kennedy2016children} and TomRoma/Child\_Speech~\citep{tomroma2024child_speech_whisper}; Chinese data from the BAAI/ChildMandarin~\citep{zhou2024childmandarin} targeting children aged 3--5. We perform rigorous filtering to exclude samples with poor signal-to-noise ratios and manually validate transcripts for accuracy.

\noindent \textbf{Older adults} Elderly speech is affected by presbyphonia, causing reduced vocal intensity, breathiness, hoarseness, tremors, and slower articulation that degrade ASR performance. We sample English speakers aged 50+ from MushanW/GLOBE\_V3~\citep{wang2024globe} and Chinese elderly speech from evan0617/seniortalk~\citep{chen2025seniortalkchineseconversationdataset}. Given that elderly speakers may exhibit confounding factors such as dialect variations, we perform manual filtering to select samples where age-related acoustic degradation is the dominant feature, minimizing the influence of other variables.

\noindent \textbf{Accent} While native accents are well-represented, global deployment requires robustness to second language accents, which introduce phonemic substitutions and stress shifts. English accented samples are drawn from MushanW/GLOBE\_V2~\citep{wang2024globe} with diverse non-native accents, while Chinese samples from TwinkStart/KeSpeech~\citep{Qundong2026ultraevalaudio} focusing on regional Mandarin varieties, excluding mutually unintelligible dialects like Cantonese. 

Note that the demographic shift subset only covers English and Chinese at this moment, as high-quality child and elderly speech resources are scarce for the other languages. We balance data quality and acquisition difficulty to ensure reproducibility of our \ourbench{} benchmark.

\begin{figure}[t]
    \centering
\includegraphics[width=\linewidth]{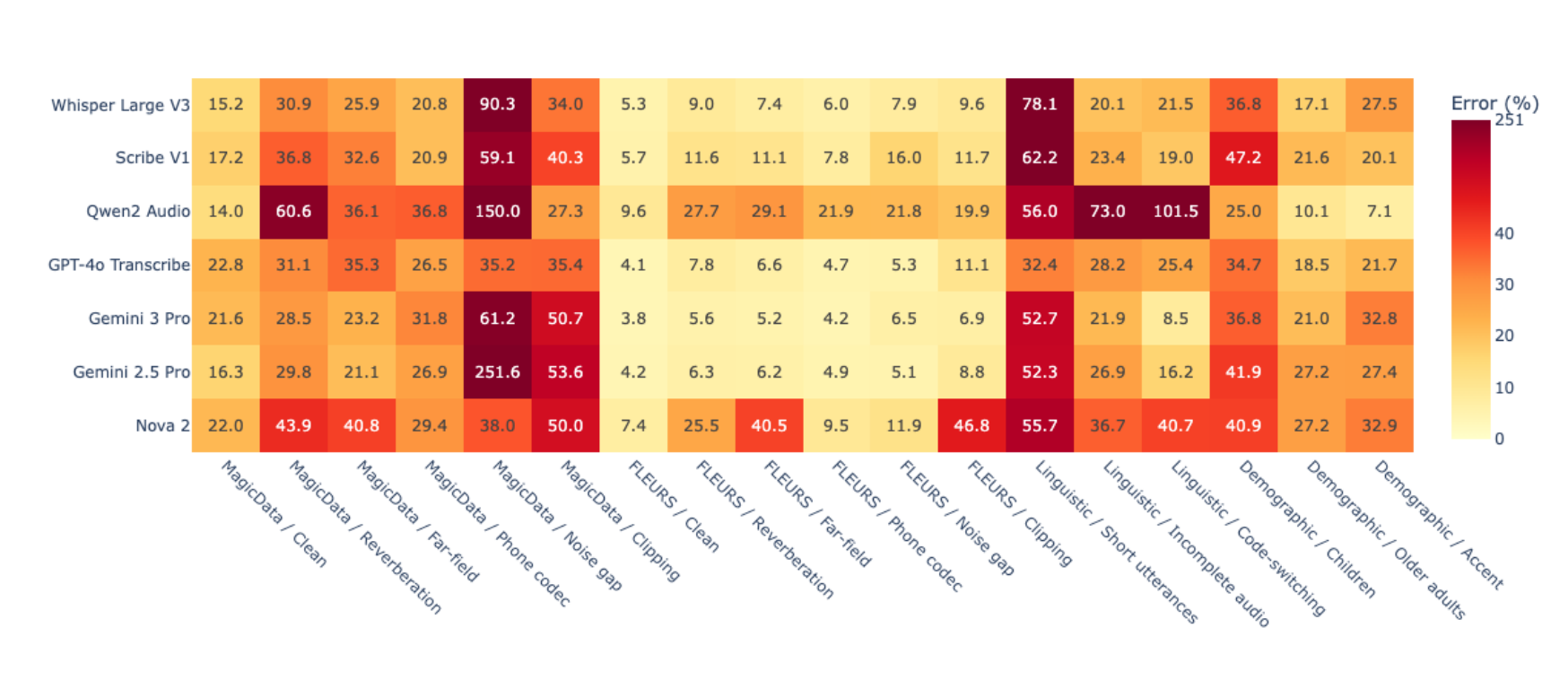}
    \vspace{-6ex}
    \caption{\textbf{Error heatmap for seven ASR models on \ourbench{}.} Each cell visualizes error rate (WER for EN and CER for CJK), with lighter colors indicating lower error. This patchy landscape reveals that ASR systems still exhibit large performance degradation and uneven robustness gaps.}
    \vspace{-2ex}
\label{fig:ood_heatmap}
\end{figure}

\vspace{-2ex}
\subsection{Linguistic diversity}
\label{subsec:linguistic_diversity}

While acoustic robustness focuses on signal quality, semantic robustness targets linguistic phenomena and structural edge cases that occur frequently in spontaneous dialogue but are systematically underrepresented in standard training corpora. In this work, we identify three specific failure modes where the model's reliance on learned probabilities becomes a liability.

\noindent \textbf{Short utterances} Real-world dialogue relies heavily on backchannels (e.g., ``hmm'' ``right''), phatic greeting  (e.g., ``how are you,'' ``what's up?'') and terse commands (e.g., ``stop,'' ``next''). These are critical for natural turn-taking and latency management in voice agents. 
However, current models suffer from such short utterance, leading to wrong transcriptions and  hallucinations in most cases.
Here we select utterances containing fewer than $6$ words (or $6$ characters for CJK languages) from YODAS~\citep{li_2023_yodas} for all four languages.

\noindent \textbf{Incomplete audio} In streaming voice applications, users are frequently cut off by aggressive voice activity detection (VAD), network latency, or interruptions. However, ASR models are typically trained on complete, well-formed sentences. When presented with a cut-off word, model may fill in a likely continuation based on language priors, producing fluent completions that were never spoken - a direct pathway to hallucination. This hallucinated completion is dangerous for agents executing API calls, where ``delete'' vs. ``delete all'' requires precise transcription of the actual audio. Given selected utterances from YODAS~\citep{li_2023_yodas}, we manually edit waveforms to truncate speech mid-sentence or mid-word while referencing the truncated transcript as the ground truth. 

\noindent \textbf{Code-switching} Code-switching is frequent in multilingual communities and is a common interaction pattern for voice agents. Most ASR models rely on an initial Language Identification token to condition generation. Code-switching breaks this ``one-utterance-one-language'' assumption. Models often force the transcribed output into a single script, resulting in phonetic transliteration errors (i.e., foreign terms are mapped to nonsensical homophones in the primary language) or simply dropping the secondary language content. Here we sample the data directly from SwitchLingua~\citep{xie_2025_SwitchLingua} and perform light-weight filtering to remove samples without rich multilingual mixes.

\vspace{-2ex}
\section{Experiments}
\label{sec:experiments}

In this work, we evaluate a total of 7 state-of-the-art ASR models on the proposed \ourbench{}, covering both proprietary and open-source models. Specifically, we include Whisper Large V3~\citep{radford_2022_whisper}, GPT-4o Transcribe~\citep{gpt4o-transcribe}, Gemini 2.5 Pro~\citep{comanici_2025_gemini25}, Gemini 3 Pro~\citep{gemini-pro-3}, Qwen2-Audio~\citep{qwen2-audio},  Nova 2~\citep{nova-2} and Scribe V1~\citep{scribe-v1}.  Details of inference protocol are included in Appendix~\ref{sec:inference_setting}. Due to space constraints, results in the main text are presented in aggregated form to facilitate cross-condition analysis; full per-model breakdowns for each subset and language are provided in Appendix~\ref{sec:detailed_results}.

\subsection{Multilingual ASR is not solved}
\label{subsec:not_solved}

To have an overall understanding of models' ASR performance on \ourbench{}, we conduct a systematic evaluation and present the general results in Figure~\ref{fig:ood_heatmap}. Each cell aggregates error across available languages. It reveals a patchy landscape where each model shows pockets of strong performance alongside severe failures, indicating that ASR systems still exhibit large performance degradation and uneven robustness gaps across realistic OOD conditions.

Furthermore, robustness does not uniformly transfer across environmental, semantic, and demographic shifts. For instance, Gemini 3 Pro achieves low error on FLEURS/Clean (3.8\%) but degrades sharply on MagicData/Noise gap (61.2\%) and Linguistic/Short utterances (52.7\%).
These patterns are common in Figure~\ref{fig:ood_heatmap}, making extrapolation from one setting to another unreliable, which indicates models can excel on standard benchmarks yet fail drastically under real-world conditions. This validates the importance of our benchmark in revealing weaknesses masked by aggregate metrics. We next present detailed findings across three dimensions to systematically analyze robustness of multilingual ASR.

\begin{table*}[t]
\centering
\small
\setlength{\tabcolsep}{5pt}
\renewcommand{\arraystretch}{1.05}
\caption{\textbf{Impact of environmental degradations on multilingual ASR performance.} Average error rates across seven ASR models under controlled acoustic perturbations. Results are reported as MagicData / FLEURS. $\Delta$ denotes the absolute increase in error rates relative to the clean condition. Bold highlights the largest degradation magnitude per language and dataset.}
\resizebox{0.9\textwidth}{!}{%
\begin{tabular}{l cc cc cc cc}
\toprule
\multirow{2}{*}{Perturbations} & \multicolumn{2}{c}{EN} & \multicolumn{2}{c}{ZH} & \multicolumn{2}{c}{JA} & \multicolumn{2}{c}{KO} \\
\cmidrule(lr){2-3} \cmidrule(lr){4-5} \cmidrule(lr){6-7} \cmidrule(lr){8-9}
 & WER (\%) & $\Delta$ & CER (\%) & $\Delta$ & CER (\%) & $\Delta$ & CER (\%) & $\Delta$ \\
\midrule
\textit{Original} & \textit{19.9/4.1} & --/-- & \textit{14.6/7.8} & --/-- & \textit{19.7/5.1} & --/-- & \textit{19.5/5.9} & --/-- \\
\midrule
Reverberation & 31.9/9.5 & +12.0/+5.3 & 25.7/13.0 & +11.1/+5.2 & 45.3/15.5 & +25.5/\textbf{+10.4} & 46.6/15.5 & +27.0/+9.6 \\
Far-field & 26.0/15.8 & +6.1/\textbf{+11.7} & 23.1/12.3 & +8.5/+4.5 & 33.7/13.5 & +13.9/+8.4 & 40.1/19.1 & +20.6/\textbf{+13.2} \\
Phone (G.711) & 20.5/10.4 & +0.6/+6.3 & 16.9/8.6 & +2.3/+0.8 & 29.1/6.7 & +9.4/+1.6 & 24.8/8.8 & +5.3/+2.9 \\
Phone (GSM) & 22.9/5.2 & +3.0/+1.1 & 25.0/9.4 & +10.4/+1.6 & 33.8/10.5 & +14.1/+5.4 & 47.6/7.9 & +28.0/+2.0 \\
Noise gap & 87.6/6.7 & \textbf{+67.7}/+2.5 & 24.9/13.2 & +10.3/+5.4 & 138.7/10.0 & \textbf{+118.9}/+5.0 & 140.5/12.8 & \textbf{+121.0}/+6.8 \\
Clipping & 30.6/15.6 & +10.7/+11.5 & 37.3/17.9 & \textbf{+22.7}/\textbf{+10.1} & 52.0/13.6 & +32.3/+8.5 & 46.6/18.5 & +27.0/+12.5 \\
\bottomrule
\end{tabular}
}
\label{tab:results_aug_env_combined}
\end{table*}

\subsection{Environmental degradation subset}
\label{subsec:environmental_results}

\begin{wrapfigure}[20]{r}{0.35\columnwidth}
\centering
\captionsetup{font=footnotesize,skip=2pt}

\begin{subfigure}{\linewidth}
  \centering
  \includegraphics[width=\linewidth,height=0.18\textheight,keepaspectratio]{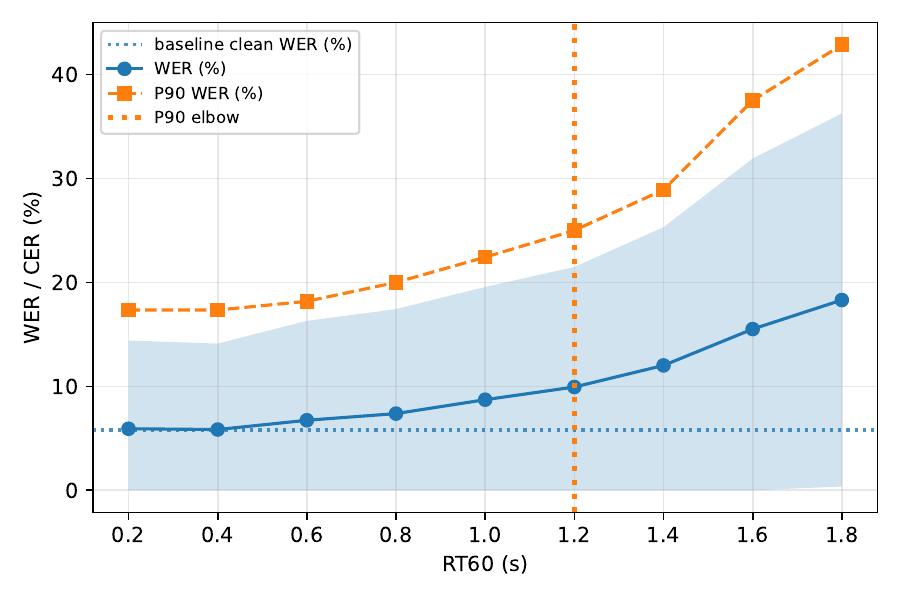}
\end{subfigure}

\begin{subfigure}{\linewidth}
  \centering
  \includegraphics[width=\linewidth,height=0.18\textheight,keepaspectratio]{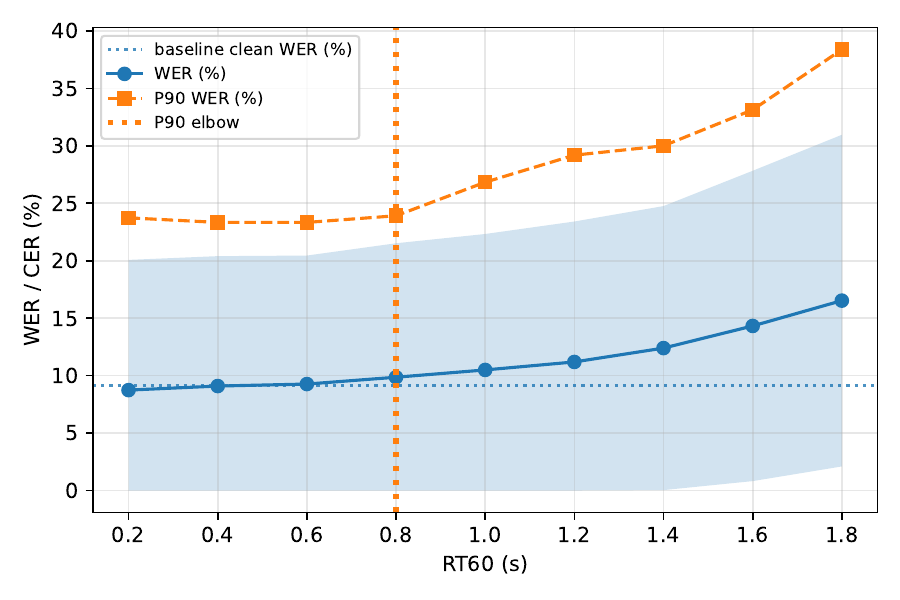}
\end{subfigure}

\captionsetup{font=small,skip=1pt} 
\caption{\textbf{ASR error dynamics under increasing reverberation} for Qwen2-Audio on FLEURS (top: English, bottom: Chinese). }
\label{fig:elbow_points}
\end{wrapfigure}

In Table~\ref{tab:results_aug_env_combined}, we report the results separately on FLEURS and MagicData to have a holistic understanding of the impact from environmental degradations on both read speech and spontaneous conversational speech. For each dataset and language, we average the resulting WER/CER across models for each perturbation type, and additionally report paired degradations as $\Delta$WER/$\Delta$CER  relative to the original (clean) condition. 

We observe that \textbf{all acoustic perturbations result in positive error increases} across both corpora, indicating that each perturbation category introduces measurable performance degradation. Overall, the average degradation is often larger on MagicData than on FLEURS, likely because conversational speech exhibits greater variability and poses more challenges than read speech. Notably, noise gap is the most detrimental perturbation for conversational speech, increasing the error rate on MagicData by +67.7\% (EN) and +10.3\% (ZH).

We also find that \textbf{degradation patterns are highly non-uniform} across languages and recording settings. For example, on MagicData, noise gap increases ZH CER by +10.3\%, yet increases JA and KO CER by +118.9\% and +121.0\%, respectively. Together, these discrepancies indicate that robustness measured in one language or recording setting can substantially mispredict behavior in another. 

In addition, we analyze performance as a function of distortion strength, using reverberation as an example with Qwen2-Audio on FLEURS, shown in Figure~\ref{fig:elbow_points}. The blue solid curve shows corpus-level WER at each distortion level, with the blue dotted line denoting the clean baseline; the shaded band indicates $\pm 1$ standard deviation of utterance-level WER. The orange dashed curve reports the P90 (90th percentile) WER, capturing tail behavior, and the vertical orange dashed line marks the P90 elbow point.
As distortion severity increases, \textbf{corpus-level WER grows gradually, while the error distribution widens substantially}: the P90 curve rises faster than the mean and variability across utterances increases. This pattern indicates the emergence of severe outliers even when average performance remains acceptable, a critical concern for voice-agent deployment where tail failures strongly affect user experience. To quantify the onset of instability, we define the P90 elbow as the distortion level at which the P90 curve exhibits accelerated growth, computed using knee-detection methods. This elbow provides a practical instability threshold for deployment decisions, such as bounding allowable distortion or triggering abstention.

\subsection{Demographic shift subset}
\label{subsec:demographic_results}

\begin{wraptable}[14]{r}{0.45\textwidth}
  \vspace{-1.2\baselineskip}
  \centering
  \small
  \caption{\textbf{ASR performance under demographic shift.} English remains relatively robust, while Chinese and child speech exhibit substantially higher error rates.}
  \label{tab:results_aug_accent}
  \setlength{\tabcolsep}{4pt}
  \renewcommand{\arraystretch}{1.05}
  \resizebox{\linewidth}{!}{%
    \begin{tabular}{l cc cc cc}
      \toprule
      & \multicolumn{2}{c}{Accent} & \multicolumn{2}{c}{Children} & \multicolumn{2}{c}{Older} \\
      \cmidrule(lr){2-3}\cmidrule(lr){4-5}\cmidrule(lr){6-7}
      Model & ZH & EN & ZH & EN & ZH & EN \\
      \midrule
      Nova 2 & 59.2 & 6.6 & 54.4 & 27.4 & 51.6 & 2.9 \\
      GPT-4o Transcribe & 40.7 & 2.6 & 39.9 & 29.4 & 36.0 & 1.1 \\
      Gemini 2.5 Pro & 49.9 & 5.0 & 58.6 & 25.1 & 52.6 & 1.8 \\
      Gemini 3 Pro & 62.5 & 3.0 & 55.3 & \textbf{18.2} & 41.4 & 0.7 \\
      Qwen2-Audio & \textbf{7.5} & 6.8 & \textbf{23.4} & 26.7 & \textbf{18.6} & 1.5 \\
      Scribe V1 & 37.9 & \textbf{2.2} & 65.1 & 29.3 & 42.3 & 0.8 \\
      Whisper Large V3 & 51.0 & 4.1 & 52.0 & 21.7 & 34.0 & \textbf{0.2} \\
      \bottomrule
    \end{tabular}
  }
\end{wraptable}

Table~\ref{tab:results_aug_accent} reports performance of seven models under demographic shift. Across models, robustness is consistently higher for English than for Chinese: WERs for Accent and Older speech in English remain in the low single digits, whereas Chinese exhibits substantially higher error rates. Notably, \textbf{child speech in English remains challenging for all models}, with the lowest observed error still at 18.2 WER, indicating a deployment-critical failure mode given the prevalence of child and family use cases. For Chinese, Qwen2-Audio shows the lowest error across all three demographic conditions, likely reflecting broader coverage in its Chinese training data.

\begin{wrapfigure}[15]{r}{0.45\textwidth} 
  \vspace{-0.5\baselineskip}
  \centering
  \includegraphics[width=\linewidth]{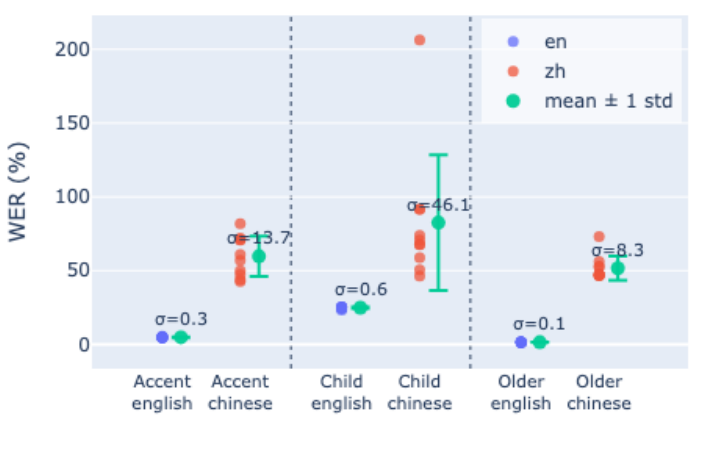}
  \captionsetup{font=small,skip=1pt} 
  
  \caption{Prompt sensitivity of Gemini 2.5 Pro on demographic subsets across ten paraphrased prompts (EN/ZH).}

  \label{fig:prompt_sensitivity}
  
\end{wrapfigure}

We further analyze prompt sensitivity in multilingual ASR by evaluating Gemini 2.5 Pro with ten paraphrased prompts on the three demographic OOD subsets in both languages. The prompts are listed in Appendix~\ref{sec:prompt_variations}. All prompts express the same instruction: transcribe the speech in the target language and output only the transcript, but differ in wording and style. For each prompt, we compute corpus-level error rates and visualize their distribution, along with mean and standard deviation, in Figure~\ref{fig:prompt_sensitivity}.

Results show that \textbf{ASR performance could be highly sensitive to prompt phrasing, particularly in Chinese}. Across Chinese subsets, the standard deviation across prompts reaches $\sigma=13.7\%$ (Accent), $\sigma=46.1\%$ (Children), and $\sigma=8.3\%$ (Older), whereas English exhibits minimal variation ($\sigma \leq 0.6\%$ across all conditions). These findings demonstrate that even for basic transcription, paraphrased instructions can materially affect model behavior, especially in non-English settings. As the optimal prompt is rarely known in advance in real-world deployments, prompt choice alone can induce substantial performance degradation. This motivates evaluating ASR systems not only by mean error under a single prompt, but also by prompt robustness, e.g., variance across a controlled prompt set as a first-class stability metric. On the other hand, the profiling methodology itself is reusable: practitioners can apply the same controlled prompt set to any new model or language to assess prompt stability before deployment.

\subsection{Linguistic diversity subset}
\label{subsec:linguistic_results}

In this section, we evaluate models across three challenging linguistic scenarios: short utterances, incomplete audio and code-switching.  
To understand hallucination behavior, we compute Hallucination Error Rate (HER)~\citep{atwany2025lost} to assess semantic-level errors beyond lexical metrics. 

\begin{table*}[t]
\centering
\small
\setlength{\tabcolsep}{6pt}
\caption{\textbf{ASR performance and hallucination behavior under linguistic diversity.} We can see that short and truncated inputs induce high error and frequent hallucinations, revealing semantic failures not captured by lexical metrics alone (EN not applicable for code-switching).}
\renewcommand{\arraystretch}{1.05}
\resizebox{0.9\textwidth}{!}{%
\begin{tabular}{llcccc|cccc}
\toprule
& & \multicolumn{4}{c}{WER/CER/MER (\%)} & \multicolumn{4}{c}{HER (\%)} \\
\cmidrule(lr){3-6} \cmidrule(lr){7-10}
Model & Category & ZH & EN & JA & KO & ZH & EN & JA & KO \\
\midrule
\rowcolor{blue!10} \cellcolor{white} Nova 2 & code-switch & 33.7 & - & 32.0 & 56.4 & 68.4 & - & 58.1 & 71.9 \\
\rowcolor{green!10} \cellcolor{white}  & short & 57.6 & 43.2 & 56.8 & 65.3 & 52.6 & 36.3 & 46.9 & 59.6 \\
\rowcolor{orange!10} \cellcolor{white}  & incomplete & 35.0 & 13.5 & 37.2 & 61.0 & 38.1 & 7.8 & 37.4 & 56.8 \\
\rowcolor{blue!10} \cellcolor{white} Gemini 2.5 Pro & code-switch & 20.7 & - & 9.8 & 18.2 & 7.0 & - & 9.4 & 19.1 \\
\rowcolor{green!10} \cellcolor{white}  & short & 40.6 & 64.4 & 48.6 & 55.6 & 30.5 & 35.4 & 28.1 & 34.1 \\
\rowcolor{orange!10} \cellcolor{white}  & incomplete & 31.9 & 15.3 & 37.1 & 23.5 & 31.6 & 10.5 & 32.8 & 11.1 \\
\rowcolor{blue!10} \cellcolor{white} Gemini 3 Pro & code-switch & 7.2 & - & 9.0 & 9.4 & 3.7 & - & 6.3 & 11.9 \\
\rowcolor{green!10} \cellcolor{white}  & short & 33.9 & 73.9 & 55.4 & 47.8 & 15.5 & 27.3 & 31.2 & 23.5 \\
\rowcolor{orange!10} \cellcolor{white}  & incomplete & 21.7 & 10.6 & 36.7 & 18.5 & 16.9 & 6.7 & 25.6 & 14.1 \\
\rowcolor{blue!10} \cellcolor{white} GPT-4o Transcribe & code-switch & 21.9 & - & 24.4 & 29.9 & 12.0 & - & 17.9 & 36.9 \\
\rowcolor{green!10} \cellcolor{white}  & short & 26.9 & 38.7 & 37.3 & 26.9 & 21.5 & 20.5 & 21.9 & 21.2 \\
\rowcolor{orange!10} \cellcolor{white}  & incomplete & 25.4 & 38.1 & 26.6 & 22.9 & 22.3 & 12.4 & 26.1 & 12.6 \\
\rowcolor{blue!10} \cellcolor{white} Qwen2-Audio & code-switch & 12.3 & - & 80.5 & 211.7 & 8.9 & - & 35.6 & 85.7 \\
\rowcolor{green!10} \cellcolor{white}  & short & 21.4 & 40.7 & 59.2 & 102.6 & 14.7 & 23.3 & 40.6 & 73.3 \\
\rowcolor{orange!10} \cellcolor{white}  & incomplete & 20.5 & 13.0 & 224.4 & 34.2 & 14.7 & 6.1 & 21.5 & 37.6 \\
\rowcolor{blue!10} \cellcolor{white} Scribe V1 & code-switch & 10.2 & - & 22.8 & 23.9 & 7.6 & - & 20.1 & 31.7 \\
\rowcolor{green!10} \cellcolor{white}  & short & 38.3 & 57.2 & 94.9 & 58.3 & 30.0 & 38.5 & 50.0 & 32.6 \\
\rowcolor{orange!10} \cellcolor{white}  & incomplete & 25.5 & 12.9 & 36.4 & 18.8 & 30.5 & 10.9 & 38.5 & 15.7 \\
\rowcolor{blue!10} \cellcolor{white} Whisper Large V3 & code-switch & 12.0 & - & 22.8 & 29.6 & 10.9 & - & 23.6 & 38.0 \\
\rowcolor{green!10} \cellcolor{white}  & short & 41.6 & 39.8 & 154.1 & 92.0 & 31.6 & 21.4 & 9.4 & 22.0 \\
\rowcolor{orange!10} \cellcolor{white}  & incomplete & 24.0 & 12.2 & 26.7 & 17.7 & 21.0 & 7.7 & 19.5 & 12.9 \\
\bottomrule
\end{tabular}
}
\label{tab:semantic_results}
\end{table*}

Detailed results can be seen in Table~\ref{tab:semantic_results}. Across all languages, \textbf{short utterances consistently induce high error rates}, reaching 38.7\%–73.9\% even in English. The reason might be threefold: (i) short segments contain limited acoustic evidence and are more sensitive to VAD errors; (ii) decoder-only models with strong language priors may over-generate plausible continuations when context is scarce; and (iii) many training pipelines downweight or remove very short clips, reducing coverage of these patterns entirely. 

We further observe \textbf{insertion-dominated auto-completion failures}, with WER/CER exceeding 100\%, indicating that models generate substantial hallucinated content rather than transcribing faithfully. For example, Qwen2-Audio reaches 102.6\% CER on KO short utterances, 211.7\% MER on KO code-switching, and 224.4\% on JA incomplete audio. These failures suggest a tendency to ``complete'' truncated or ambiguous inputs instead of producing conservative transcriptions.

Finally, \textbf{HER reveals semantic failures that lexical metrics alone obscure}. Discrepancies between WER/CER and HER highlight cases where surface-level transcription appears reasonable despite severe meaning distortion. For instance, Nova 2 on ZH code-switching exhibits 33.7\% MER but 68.4\% HER, indicating substantial semantic fabrication. Such meaning-altering hallucinations, e.g., negation introduced by a single insertion (``no I can'' $\rightarrow$ ``no I can't'') pose significant risks in high-stakes applications. Joint analysis of WER and HER therefore enables a more faithful characterization of ASR reliability by distinguishing benign lexical errors from critical semantic failures. This joint evaluation protocol can be readily applied to any ASR system to surface semantic risks that WER alone would miss.

Overall, incomplete audio is relatively manageable for EN, while code-switching is substantially harder for JA/KO mixed with English. Another observation is, proprietary models tend to perform better (more robust) than open-source models on our linguistic diversity subset, e.g., occurrences where error rates beyond 80 happen mostly in Qwen2-Audio and Whisper Large V3.

\section{Further Discussion}
\label{sec:discussion}

\noindent\textbf{Is \ourbench{} intrinsically difficult for humans?}
Although \ourbench{} induces severe failures in state-of-the-art ASR systems, the underlying speech remains largely intelligible to human listeners. \textit{We conducted a human evaluation in which samples were reviewed in randomized and anonymized order by independent annotators. The resulting average error rate was 4.7\%, consistent with established estimates of human-level transcription performance.} This gap confirms that \ourbench{} does not derive its difficulty from ambiguity or poor signal quality, but rather exposes modeling limitations under realistic long-tail conditions. The disparity between human and model performance highlights substantial headroom for improving robustness in deployed ASR systems.

\noindent\textbf{Why real speech sources matter for robustness evaluation?}
A central design choice of \ourbench{} is that all source audio originates from real human speech, rather than being generated by text-to-speech systems. To assess the impact of this choice, \textit{we compared real child speech with synthetic counterparts generated from identical transcripts using Qwen3-TTS~\citep{hu2026qwen3ttstechnicalreport}. Take Whisper Large V3 as an example, it achieved near-ceiling performance on synthetic audio (3.7\%), but error rates increased dramatically on real child speech (21.7\%) for English (shown in table \ref{tab:results_aug_accent}).} Qualitative inspection reveals that synthetic samples capture coarse acoustic cues (e.g., pitch) but fail to reproduce authentic paralinguistic phenomena such as hesitations and unstable articulation. This discrepancy suggests that evaluations relying on synthetic data can underestimate failure rates. Real speech remains essential for revealing robustness gaps that directly affect voice-agent reliability.

\noindent\textbf{What is the ``ground truth'' for multilingual ASR?}
Although ASR is often treated as a well-defined transcription task, its notion of ground truth is inherently use-case and culture dependent. Decisions such as whether to preserve filler words or partial utterances vary across languages and conversational norms, and can materially affect downstream interpretation. In some settings, these phenomena convey pragmatic meaning, while in others they are routinely normalized. While \ourbench{} adopts a fixed transcription target for consistency, our observations highlight the need for multilingual benchmarks that account for culturally specific transcription norms and evaluate how different normalization choices impact robustness, hallucination behavior, and downstream utility.

\noindent \textbf{Is ASR obsolete in the era of speech-to-speech systems?} Recent progress in large multimodal and S2S models has motivated the view that explicit transcription may become unnecessary, as end-to-end systems can directly operate on acoustic representations while preserving paralinguistic cues. Indeed, modern voice agents can often sustain fluent conversations even when retrospective transcripts contain recognition errors or hallucinated phrases. However, our results argue that this does not diminish the importance of ASR; rather, it reframes its role. Explicit ASR provides a transparent, auditable, and inspectable interface that is critical for debugging, compliance, retrieval, indexing, and structured tool use. Moreover, while S2S systems may tolerate minor errors in benign conditions, our findings on severe hallucinations under OOD inputs suggest that uninterpretable end-to-end failures may be harder to detect and correct. From this perspective, robust ASR should be viewed not as a legacy component, but as a stabilizing input layer and safety guardrail for next-generation voice agents. Future work should study hybrid architectures that dynamically combine explicit transcription with audio-native reasoning, rather than treating them as mutually exclusive.

\section{Conclusion and future work}
\label{sec:conclusion}

This work introduces \ourbench{}, a multilingual benchmark designed to stress-test ASR systems under a diverse set of OOD conditions spanning acoustic environments (\emph{where}), demographic characteristics (\emph{who}), and linguistic phenomena (\emph{what}). Across all evaluated models, our results reveal a fragmented robustness landscape: strong performance on in-domain benchmarks does not reliably transfer across domains, demographics, or interaction settings, and failure modes often manifest as severe semantic distortions rather than gradual degradation.
These findings carry broader implications beyond ASR as a standalone task since voice interfaces and conversational agents become an increasingly prominent mode of human--AI interaction.

Due to the scope of constructing a multilingual benchmark from real human speech sources and the resources required for large-scale evaluation, \ourbench{} naturally has limitations that point to promising future directions:

\noindent \textbf{From diagnosis to mitigation.} Our current work focuses on identifying and characterizing failure modes, including hallucination behavior, rather than resolving them. The factor-isolated structure of \ourbench{} directly reveals which OOD conditions cause the largest degradation for each model and language, providing a natural starting point for targeted data augmentation, fine-tuning, or adaptation strategies. In particular, our hallucination analysis motivates the development of hallucination-aware decoding and abstention mechanisms that withhold transcription when confidence is low.

\noindent \textbf{Broader language, condition, and sample coverage.} The current benchmark covers four languages and a defined set of OOD factors, leaving out low-resource languages and conditions such as multi-speaker overlap and real-time streaming artifacts. Additionally, certain subsets, particularly the demographic split, have limited sample sizes due to the scarcity of publicly available real speech data for specific populations. As a diagnostic benchmark designed to expose failure modes and identify broad degradation patterns, these sizes are sufficient for our analytical goals, but expanding both language coverage and per-condition sample sizes would further strengthen the diagnostic coverage.

\noindent \textbf{Finer-grained reporting and broader model coverage.} Due to space constraints, results in the main text are presented in aggregated form to support cross-condition analysis and narrative clarity; full per-model breakdowns are provided in the appendix. We also welcome future work that evaluates additional models on \ourbench{} to broaden the comparative landscape.

\noindent More broadly, \ourbench{} highlights the need for benchmarks grounded in real human speech sources and realistic usage patterns, as synthetic or narrowly curated evaluations risk obscuring failure modes that matter most in deployment. We hope this work serves as a foundation for developing ASR systems that are not only accurate, but dependable in real-world voice agents.



\newpage
\bibliography{asrbench}
\bibliographystyle{iclr2026_conference}

\newpage
\appendix

\section{Unified inference protocol}
\label{sec:inference_setting}

All systems are evaluated under a unified inference protocol unless specified otherwise.  We evaluate each model on all subsets listed in Table~\ref{tab:benchmark-construction}, and report performance independently for each factor. We report corpus-level WER for English (EN) and CER for Chinese/Japanese/Korean (ZH/JA/KO). 
For the code-switching subset, we report Mixed Error Rate (MER): each transcript is tokenized into a mixed sequence where Latin/English spans are word-tokenized (after normalization) and CJK scripts are character-tokenized, and we compute WER over the mixed token stream at the corpus level. Hyperparameter settings can be seen in Table\ref{tab:asr_inference_settings}.

\begin{table}[h]
\centering
\small
\setlength{\tabcolsep}{7pt}
\renewcommand{\arraystretch}{1.15}
\begin{tabular}{l c}
\toprule
\textbf{Inference setting} & \textbf{Value} \\
\midrule
Temperature & 0.2 \\
Top-$p$ & 0.9 \\
Max new tokens & 2048 \\
Language conditioning & Prompt (\texttt{\{language\_name\}}) and/or SDK language code \\
\bottomrule
\end{tabular}
\caption{
\textbf{Unified inference setting for ASR benchmarking.} Audio inputs are resampled to 16kHz. The default instruction is:
\texttt{`Please transcribe the audio in \{language\_name\}. Do not add any additional text that is not in the speech content.'}}
\label{tab:asr_inference_settings}
\end{table}

\section{Accent distribution}
\label{sec:accent_distribution}

\begin{figure}[h]
    \centering
    \begin{subfigure}{0.40\textwidth}
        \centering
        \includegraphics[width=\linewidth]{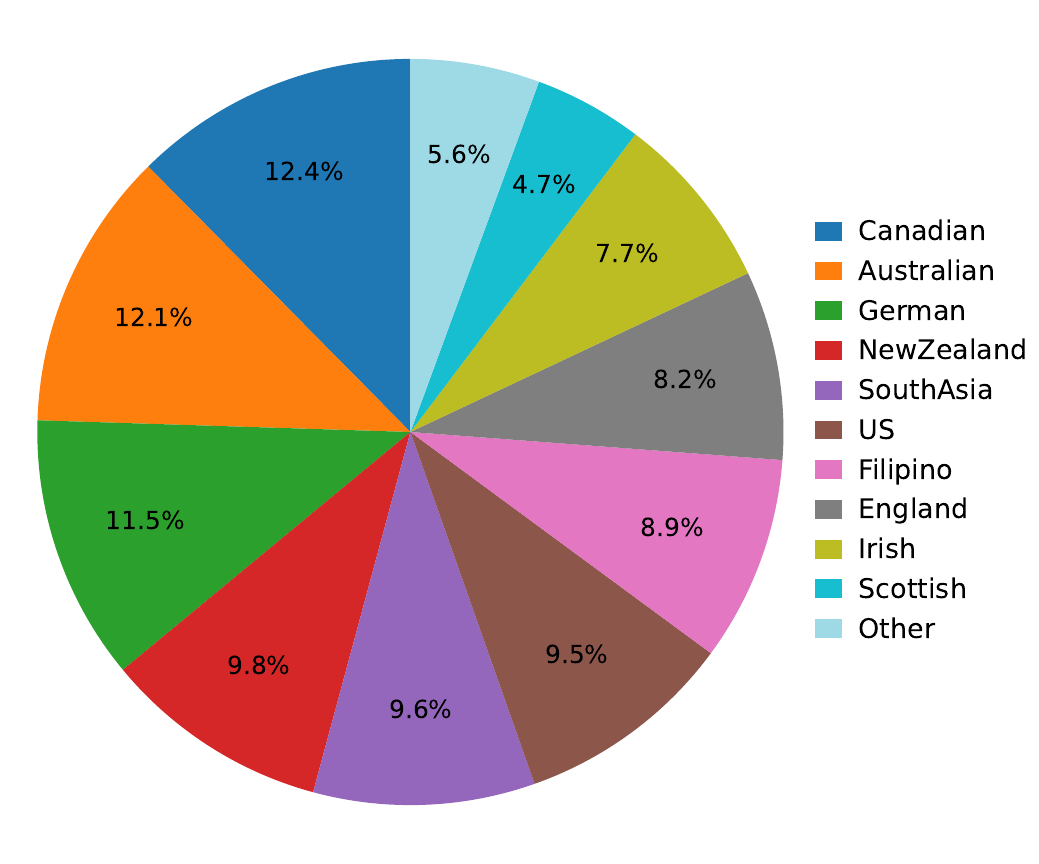}
        \label{fig:en_accent}
    \end{subfigure}
    \vspace{-2ex}
    \hfill
    \begin{subfigure}{0.40\textwidth}
        \centering
        \includegraphics[width=\linewidth]{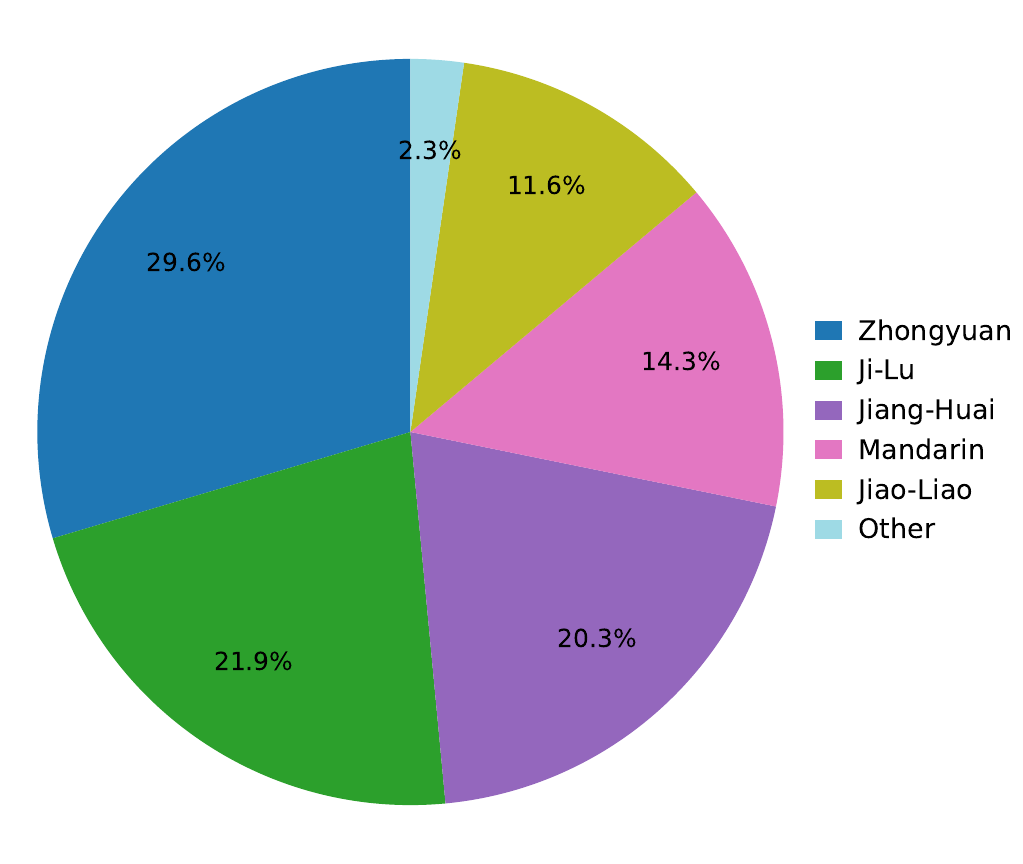}
        \label{fig:zh_accent}
    \end{subfigure}

    \caption{Accent Distribution in \ourbench{}. Left: English. Right: Chinese. }
    \label{fig:accent_distribution}
\end{figure}

The accent distribution in \ourbench{} is visualized in Figure~\ref{fig:accent_distribution}. For English (Figure~\ref{fig:accent_distribution} left), the dataset encompasses a diverse range of accents including Canadian (12.4\%), Australian (12.1\%), German (11.5\%), etc. This distribution ensures comprehensive coverage of non-native English accents. For Chinese (Figure~\ref{fig:accent_distribution} right), the dataset focuses on regional Mandarin varieties with representation from Zhongyuan (29.6\%), Ji Lu (21.9\%), Jiang Huai (20.3\%), etc, capturing the phonological diversity across different Mandarin-speaking regions while maintaining mutual intelligibility.

\section{Curation pipeline details}
\label{sec:curation_details}

We describe each stage of the curation pipeline introduced in \S\ref{subsec:curation}:

\noindent \textbf{DC (Data Collection).} We source raw audio from publicly available speech corpora with verified transcripts across our target languages. Source datasets for each subcategory are listed in Table~\ref{tab:data_sources}.

\noindent \textbf{SF (Speaker Filtering).} For demographic subsets, we verify speaker metadata (age, accent, native language) against dataset annotations and discard samples with ambiguous or missing labels. For older adults, we filter to select samples where age-related acoustic degradation is the dominant feature, minimizing confounding factors such as dialect variation.

\noindent \textbf{QF (Quality Filtering).} We discard unintelligible or corrupted recordings and remove samples with poor signal-to-noise ratios. For children speech, we perform rigorous filtering to exclude low-SNR samples. For code-switching, we remove samples without substantive multilingual mixing.

\noindent \textbf{NR (Audio Normalization).} All audio is resampled to 16\,kHz mono with loudness normalization to ensure consistent input conditions across sources.

\noindent \textbf{AA (Acoustic Augmentation).} For the environmental degradation subset, we apply five controlled, transcript-preserving perturbations (reverberation, far-field, phone codec, noise gap, clipping) at multiple calibrated severity levels, as detailed in \S\ref{subsec:environmental_degradation}.

\noindent \textbf{MT (Manual Truncation \& Transcript Alignment).} For incomplete audio, we manually edit waveforms to truncate speech mid-sentence or mid-word, and use the truncated transcript as the ground truth, as described in \S\ref{subsec:linguistic_diversity}.

\noindent \textbf{MV (Manual Verification).} We manually validate transcript correctness across all subsets. For children and older adult speech, transcripts are manually reviewed for accuracy.

\section{Data sources}
\label{sec:data_sources}

Table~\ref{tab:data_sources} lists the source datasets used for each subcategory of \ourbench{}.

\begin{table}[h]
\centering
\small
\setlength{\tabcolsep}{5pt}
\renewcommand{\arraystretch}{1.15}
\caption{\textbf{Source datasets for each \ourbench{} subcategory.}}
\label{tab:data_sources}
\begin{tabular}{l l}
\toprule
\textbf{Subcategory} & \textbf{Sources} \\
\midrule
Environmental degradation & FLEURS, MagicData \\
Children & Zenodo, Child\_Speech, ChildMandarin \\
Older adults & GLOBE\_V3, seniortalk \\
Accent & GLOBE\_V2, KeSpeech \\
Short utterances & YODAS \\
Incomplete audio & YODAS \\
Code-switching & SwitchLingua \\
\bottomrule
\end{tabular}
\end{table}

\section{Prompt sensitivity}
\label{sec:prompt_variations}

We run prompt-sensitivity experiments in two languages (English and Chinese) using Gemini 2.5 Pro to measure how much the model’s transcripts change when the instruction wording changes. We evaluate on demographic slices (Accent, Children, Older Adult). For fairness, we express the same transcription request using 10 paraphrased prompt variants in each language. For each audio sample, we test all 10 variants in the sample’s original language. The English prompt variants are shown in Table~\ref{tab:prompt_variants_en}.

\vspace{2ex}

\begin{table*}[h]
\centering
\small
\setlength{\tabcolsep}{6pt}
\renewcommand{\arraystretch}{1.15}
\caption{\textbf{English prompt variants used for prompt-sensitivity evaluation.} All prompts share the same intent---\emph{transcribe the audio into \{english\} and output only the transcript}---but differ in wording and style.}
\begin{tabular}{cp{13cm}}
\toprule
\textbf{ID} & \textbf{English prompt variant} \\
\midrule
p01 & English transcription only. Return the spoken words as text---nothing else. \\
p02 & Task: produce a verbatim english transcript of the audio. Output transcript text only; omit all other content. \\
p03 & Perform speech-to-text for this audio in english. Respond with plain transcript text only (no labels or commentary). \\
p04 & Please write the english transcript of what you hear. Reply with only the transcription. \\
p05 & Create the most faithful english transcript possible. Output ONLY the transcript (no summaries, notes, or formatting). \\
p06 & In english, transcribe the audio. Provide only the transcript. \\
p07 & You are an ASR engine. Emit the english transcript as raw text only; do not add headers, punctuation notes, or extra lines. \\
p08 & Return exactly one thing: the english transcription of the audio. No preamble, no JSON, no quotes---just the transcript. \\
p09 & Write the spoken content as a english transcript. Do not include timestamps, speaker tags, explanations, or any non-transcript text. \\
p10 & Provide the english transcript of the audio. Your entire response should be the transcript text and nothing more. \\
\bottomrule
\end{tabular}
\label{tab:prompt_variants_en}
\end{table*}

\section{Qualitative failure patterns}
\label{sec:failure_patterns}

Table~\ref{tab:qualitative_fail_cases} shows representative errors made by different models on the English subset, across several dataset slices. We highlight the incorrect parts of the model prediction in \err{bold}. We observe several recurring failure types:

\begin{itemize}
    \item \textbf{Non-transcription outputs} (e.g., producing tokens such as \mono{[noise]} instead of words)
    \item \textbf{Full hallucinations} (e.g., ``ah yeah'' $\rightarrow$ ``I'm with breast''; ``who identified'' $\rightarrow$ ``I don't know if I can.'')
    \item \textbf{Auto-Completion} beyond the audio (e.g., continuing ``so Putin took the'' with invented content)
    \item \textbf{Refusals} (e.g., responding with apologies or capability disclaimers instead of transcribing)
    \item \textbf{Phonetically similar substitutions}(e.g. `searching'' $\rightarrow$ ``shouting'')
\end{itemize} 
Reporting these qualitative errors is important because it reveals failure modes that are not well captured by WER/CER. These errors can be semantically plausible yet not present in the audio, which can introduce significant risks for downstream systems that rely on accurate transcripts.

\begin{table*}[!htbp]
\centering
\small
\setlength{\tabcolsep}{5pt}
\renewcommand{\arraystretch}{1.18}
\caption{\textbf{Representative English failure cases} from \ourbench{}. Rows are grouped by OOD dimension. We highlight the erroneous or safety-relevant portion in \err{bold}}
\begin{tabular}{l  p{5.25cm} p{5.25cm}}

\toprule
 \textbf{Subset} & \textbf{Ground Truth} & \textbf{Model Prediction} \\
\midrule

\multicolumn{3}{l}{\textbf{Environmental degradation}} \\
\midrule
 Noise gap  &
``a car bomb detonated at police headquarters in gaziantep turkey yesterday morning killed two police officers and injured more than twenty other people'' &
``a car bomb detonated at police headquarters in geyve tepe turkey yesterday morning killing two police officers and injuring more than a \err{hundred} other people'' \\

phone codec & 
``the center of tibetan meditation is the deity yoga through the visualization of various deities the energy channels are cleaned the chakras are activated and the enlightenment consciousness is created'' &
\textbf{``\mono{[noise]} \mono{[sigh]} \mono{[sigh]} \mono{[sigh]} \mono{[sigh]} \mono{[sigh]} \mono{[sigh]} \mono{[sigh]}} \ldots'' \\

clipping & 
``ah yeah.'' &
``\err{I'm with breast}.'' \\

\midrule
\multicolumn{3}{l}{\textbf{Demographic shift}} \\
\midrule
child & 
``he was searching for him everywhere'' &
``\err{they were shouting} for him everywhere'' \\

\midrule
\multicolumn{3}{l}{\textbf{Linguistic diversity}} \\
\midrule
incomplete audio & 
``who identified'' &
``\err{I don't know if I can.}'' \\

short utterance & 
``Yeah'' &
``\err{I'm sorry, but I cannot listen to audio files.} I can only process and generate text \ldots'' \\

short utterance & 
``so Putin took the'' &
``So Putin took the \err{measure of the West and he decided that he could handle whatever we threw at him}.'' \\

\bottomrule
\end{tabular}
\label{tab:qualitative_fail_cases}
\end{table*}

\section{Detailed per-model results}
\label{sec:detailed_results}

This appendix provides full per-model results for each subset and language, complementing the aggregated tables in the main text.

\begin{table*}[h]
\centering
\small
\setlength{\tabcolsep}{6pt}
\renewcommand{\arraystretch}{1.05}
\caption{\textbf{Nova 2} --- Environmental degradation on FLEURS. $\Delta$: absolute change relative to original.}
\label{tab:aug_wer_delta_nova2_fleurs}
\begin{tabular}{cc cc cc cc cc}
\toprule
\multicolumn{2}{c}{Augmentation}
& \multicolumn{2}{c}{ZH}
& \multicolumn{2}{c}{EN}
& \multicolumn{2}{c}{JA}
& \multicolumn{2}{c}{KO} \\
\cmidrule(lr){1-2} \cmidrule(lr){3-4} \cmidrule(lr){5-6} \cmidrule(lr){7-8} \cmidrule(lr){9-10}
Perturbation & Method
& CER (\%) & $\Delta$
& WER (\%) & $\Delta$
& CER (\%) & $\Delta$
& CER (\%) & $\Delta$ \\
\midrule
\textit{Original} & -- & 10.1 & -- & 6.0 & -- & 7.0 & -- & 6.5 & -- \\
Clipping & -- & 42.6 & \textbf{+32.5} & 60.4 & +54.4 & 32.0 & \textbf{+25.0} & 51.9 & \textbf{+45.4} \\
Far-field & -- & 25.6 & +15.5 & 71.0 & \textbf{+65.0} & 19.5 & +12.5 & 45.9 & +39.4 \\
\multirow{2}{*}{Phone} & G.711 & 12.6 & +2.5 & 6.6 & +0.6 & 8.8 & +1.8 & 6.6 & +0.1 \\
& GSM & 14.5 & +4.4 & 7.9 & +1.9 & 11.6 & +4.6 & 7.7 & +1.2 \\
Noise gap & -- & 16.8 & +6.7 & 8.2 & +2.2 & 11.9 & +4.9 & 10.7 & +4.2 \\
Reverberation & -- & 23.3 & +13.2 & 31.0 & +24.9 & 17.1 & +10.1 & 30.8 & +24.3 \\
\bottomrule
\end{tabular}
\end{table*}

\begin{table*}[h]
\centering
\small
\setlength{\tabcolsep}{6pt}
\renewcommand{\arraystretch}{1.05}
\caption{\textbf{Gemini 2.5 Pro} --- Environmental degradation on FLEURS. $\Delta$: absolute change relative to original.}
\label{tab:aug_wer_delta_gemini25_fleurs}
\begin{tabular}{cc cc cc cc cc}
\toprule
\multicolumn{2}{c}{Augmentation}
& \multicolumn{2}{c}{ZH}
& \multicolumn{2}{c}{EN}
& \multicolumn{2}{c}{JA}
& \multicolumn{2}{c}{KO} \\
\cmidrule(lr){1-2} \cmidrule(lr){3-4} \cmidrule(lr){5-6} \cmidrule(lr){7-8} \cmidrule(lr){9-10}
Perturbation & Method
& CER (\%) & $\Delta$
& WER (\%) & $\Delta$
& CER (\%) & $\Delta$
& CER (\%) & $\Delta$ \\
\midrule
\textit{Original} & -- & 6.7 & -- & 3.6 & -- & 2.7 & -- & 4.0 & -- \\
Clipping & -- & 13.6 & \textbf{+6.9} & 7.9 & \textbf{+4.3} & 6.3 & +3.6 & 7.4 & \textbf{+3.4} \\
Far-field & -- & 9.6 & +2.9 & 6.5 & +2.9 & 4.2 & +1.5 & 4.5 & +0.6 \\
\multirow{2}{*}{Phone} & G.711 & 7.0 & +0.3 & 3.5 & -0.1 & 3.2 & +0.5 & 4.0 & +0.0 \\
& GSM & 7.9 & +1.2 & 4.2 & +0.6 & 5.4 & +2.7 & 4.1 & +0.1 \\
Noise gap & -- & 8.2 & +1.5 & 4.2 & +0.6 & 3.5 & +0.8 & 4.4 & +0.4 \\
Reverberation & -- & 9.2 & +2.5 & 4.7 & +1.1 & 6.4 & \textbf{+3.7} & 5.2 & +1.2 \\
\bottomrule
\end{tabular}
\end{table*}

\begin{table*}[h]
\centering
\small
\setlength{\tabcolsep}{6pt}
\renewcommand{\arraystretch}{1.05}
\caption{\textbf{Gemini 3 Pro} --- Environmental degradation on FLEURS. $\Delta$: absolute change relative to original.}
\label{tab:aug_wer_delta_gemini3_fleurs}
\begin{tabular}{cc cc cc cc cc}
\toprule
\multicolumn{2}{c}{Augmentation}
& \multicolumn{2}{c}{ZH}
& \multicolumn{2}{c}{EN}
& \multicolumn{2}{c}{JA}
& \multicolumn{2}{c}{KO} \\
\cmidrule(lr){1-2} \cmidrule(lr){3-4} \cmidrule(lr){5-6} \cmidrule(lr){7-8} \cmidrule(lr){9-10}
Perturbation & Method
& CER (\%) & $\Delta$
& WER (\%) & $\Delta$
& CER (\%) & $\Delta$
& CER (\%) & $\Delta$ \\
\midrule
\textit{Original} & -- & 6.1 & -- & 2.8 & -- & 2.7 & -- & 3.8 & -- \\
Clipping & -- & 10.8 & +4.8 & 5.7 & \textbf{+2.9} & 5.8 & +3.1 & 5.4 & \textbf{+1.6} \\
Far-field & -- & 7.7 & +1.6 & 5.1 & +2.3 & 3.7 & +1.1 & 4.4 & +0.6 \\
\multirow{2}{*}{Phone} & G.711 & 6.2 & +0.1 & 2.9 & +0.1 & 2.8 & +0.2 & 3.8 & +0.0 \\
& GSM & 6.8 & +0.8 & 3.5 & +0.6 & 4.2 & +1.6 & 3.8 & +0.1 \\
Noise gap & -- & 11.9 & \textbf{+5.9} & 3.1 & +0.3 & 6.6 & \textbf{+4.0} & 4.2 & +0.5 \\
Reverberation & -- & 7.9 & +1.9 & 4.1 & +1.2 & 5.6 & +2.9 & 4.9 & +1.1 \\
\bottomrule
\end{tabular}
\end{table*}

\begin{table*}[h]
\centering
\small
\setlength{\tabcolsep}{6pt}
\renewcommand{\arraystretch}{1.05}
\caption{\textbf{GPT-4o Transcribe} --- Environmental degradation on FLEURS. $\Delta$: absolute change relative to original.}
\label{tab:aug_wer_delta_gpt4o_fleurs}
\begin{tabular}{cc cc cc cc cc}
\toprule
\multicolumn{2}{c}{Augmentation}
& \multicolumn{2}{c}{ZH}
& \multicolumn{2}{c}{EN}
& \multicolumn{2}{c}{JA}
& \multicolumn{2}{c}{KO} \\
\cmidrule(lr){1-2} \cmidrule(lr){3-4} \cmidrule(lr){5-6} \cmidrule(lr){7-8} \cmidrule(lr){9-10}
Perturbation & Method
& CER (\%) & $\Delta$
& WER (\%) & $\Delta$
& CER (\%) & $\Delta$
& CER (\%) & $\Delta$ \\
\midrule
\textit{Original} & -- & 6.4 & -- & 2.8 & -- & 3.0 & -- & 4.0 & -- \\
Clipping & -- & 17.4 & \textbf{+11.0} & 8.8 & \textbf{+5.9} & 7.9 & +4.9 & 10.2 & \textbf{+6.2} \\
Far-field & -- & 8.3 & +1.9 & 7.5 & +4.7 & 5.5 & +2.5 & 5.1 & +1.1 \\
\multirow{2}{*}{Phone} & G.711 & 6.7 & +0.3 & 2.9 & +0.0 & 3.7 & +0.7 & 4.0 & +0.1 \\
& GSM & 7.3 & +0.9 & 3.1 & +0.3 & 5.8 & +2.8 & 4.3 & +0.3 \\
Noise gap & -- & 7.6 & +1.2 & 4.5 & +1.6 & 4.5 & +1.5 & 4.9 & +0.9 \\
Reverberation & -- & 10.0 & +3.6 & 5.1 & +2.3 & 9.7 & \textbf{+6.7} & 6.3 & +2.4 \\
\bottomrule
\end{tabular}
\end{table*}

\begin{table*}[h]
\centering
\small
\setlength{\tabcolsep}{6pt}
\renewcommand{\arraystretch}{1.05}
\caption{\textbf{Qwen2-Audio} --- Environmental degradation on FLEURS. $\Delta$: absolute change relative to original.}
\label{tab:aug_wer_delta_qwen2audio_fleurs}
\begin{tabular}{cc cc cc cc cc}
\toprule
\multicolumn{2}{c}{Augmentation}
& \multicolumn{2}{c}{ZH}
& \multicolumn{2}{c}{EN}
& \multicolumn{2}{c}{JA}
& \multicolumn{2}{c}{KO} \\
\cmidrule(lr){1-2} \cmidrule(lr){3-4} \cmidrule(lr){5-6} \cmidrule(lr){7-8} \cmidrule(lr){9-10}
Perturbation & Method
& CER (\%) & $\Delta$
& WER (\%) & $\Delta$
& CER (\%) & $\Delta$
& CER (\%) & $\Delta$ \\
\midrule
\textit{Original} & -- & 9.1 & -- & 5.8 & -- & 10.9 & -- & 12.7 & -- \\
Clipping & -- & 11.6 & +2.5 & 12.6 & +6.8 & 19.6 & +8.7 & 35.8 & +23.1 \\
Far-field & -- & 10.1 & +1.0 & 8.9 & +4.1 & 40.2 & +29.3 & 57.0 & +44.3 \\
\multirow{2}{*}{Phone} & G.711 & 8.8 & -0.3 & 48.3 & \textbf{+42.5} & 15.3 & +4.4 & 30.9 & +18.2 \\
& GSM & 9.4 & +0.3 & 8.8 & +3.0 & 30.4 & +29.5 & 23.3 & +10.6 \\
Noise gap & -- & 16.6 & +7.5 & 9.8 & +4.0 & 20.7 & +9.8 & 40.0 & +27.3 \\
Reverberation & -- & 11.1 & +2.0 & 9.5 & +3.7 & 46.2 & +35.3 & 44.1 & +21.4 \\
\bottomrule
\end{tabular}
\end{table*}

\begin{table*}[h]
\centering
\small
\setlength{\tabcolsep}{6pt}
\renewcommand{\arraystretch}{1.05}
\caption{\textbf{Scribe V1} --- Environmental degradation on FLEURS. $\Delta$: absolute change relative to original.}
\label{tab:aug_wer_delta_scribev1_fleurs}
\begin{tabular}{cc cc cc cc cc}
\toprule
\multicolumn{2}{c}{Augmentation}
& \multicolumn{2}{c}{ZH}
& \multicolumn{2}{c}{EN}
& \multicolumn{2}{c}{JA}
& \multicolumn{2}{c}{KO} \\
\cmidrule(lr){1-2} \cmidrule(lr){3-4} \cmidrule(lr){5-6} \cmidrule(lr){7-8} \cmidrule(lr){9-10}
Perturbation & Method
& CER (\%) & $\Delta$
& WER (\%) & $\Delta$
& CER (\%) & $\Delta$
& CER (\%) & $\Delta$ \\
\midrule
\textit{Original} & -- & 8.7 & -- & 3.6 & -- & 4.8 & -- & 5.6 & -- \\
Clipping & -- & 15.7 & +7.1 & 5.6 & +2.0 & 15.9 & \textbf{+11.1} & 9.6 & +4.1 \\
Far-field & -- & 14.0 & +5.4 & 5.5 & +1.9 & 14.5 & +9.7 & 10.6 & +5.1 \\
\multirow{2}{*}{Phone} & G.711 & 11.0 & +2.4 & 4.2 & +0.6 & 8.0 & +3.2 & 6.4 & +0.8 \\
& GSM & 11.5 & +2.8 & 4.3 & +0.7 & 10.1 & +5.3 & 6.9 & +1.3 \\
Noise gap & -- & 20.3 & \textbf{+11.6} & 11.3 & \textbf{+7.7} & 14.8 & +10.0 & 17.8 & \textbf{+12.3} \\
Reverberation & -- & 16.5 & +7.8 & 5.1 & +1.5 & 14.7 & +9.9 & 9.9 & +4.4 \\
\bottomrule
\end{tabular}
\end{table*}

\begin{table*}[h]
\centering
\small
\setlength{\tabcolsep}{6pt}
\renewcommand{\arraystretch}{1.05}
\caption{\textbf{Whisper Large V3} --- Environmental degradation on FLEURS. $\Delta$: absolute change relative to original.}
\label{tab:aug_wer_delta_whisperv3_fleurs}
\begin{tabular}{cc cc cc cc cc}
\toprule
\multicolumn{2}{c}{Augmentation}
& \multicolumn{2}{c}{ZH}
& \multicolumn{2}{c}{EN}
& \multicolumn{2}{c}{JA}
& \multicolumn{2}{c}{KO} \\
\cmidrule(lr){1-2} \cmidrule(lr){3-4} \cmidrule(lr){5-6} \cmidrule(lr){7-8} \cmidrule(lr){9-10}
Perturbation & Method
& CER (\%) & $\Delta$
& WER (\%) & $\Delta$
& CER (\%) & $\Delta$
& CER (\%) & $\Delta$ \\
\midrule
\textit{Original} & -- & 7.5 & -- & 4.2 & -- & 4.6 & -- & 5.0 & -- \\
Clipping & -- & 13.5 & \textbf{+5.9} & 8.5 & \textbf{+4.3} & 7.5 & +3.0 & 9.0 & \textbf{+4.0} \\
Far-field & -- & 10.8 & +3.3 & 6.3 & +2.0 & 6.6 & +2.1 & 6.1 & +1.1 \\
\multirow{2}{*}{Phone} & G.711 & 7.8 & +0.3 & 4.8 & +0.6 & 5.0 & +0.5 & 5.9 & +0.9 \\
& GSM & 8.7 & +1.1 & 4.9 & +0.7 & 5.6 & +1.1 & 5.2 & +0.3 \\
Noise gap & -- & 10.8 & +3.3 & 5.5 & +1.3 & 8.3 & +3.8 & 7.1 & +2.2 \\
Reverberation & -- & 13.1 & +5.5 & 6.8 & +2.6 & 9.0 & \textbf{+4.5} & 7.1 & +2.2 \\
\bottomrule
\end{tabular}
\end{table*}

\end{document}